%% file: main.tex
\documentclass{article}
\usepackage{spconf,amsmath,graphicx}
\usepackage{array}
\usepackage{hyperref}
\hypersetup{colorlinks}
\newcolumntype{C}[1]{>{\centering\arraybackslash}p{#1}}

\title{Building Lane-Level Maps from Aerial Images}

\name{Jiawei Yao$^{1*}$\thanks{$*$ These authors contributed equally to this work.} \quad Xiaochao Pan$^{2*}$ \quad Tong Wu $^1$ \quad Xiaofeng Zhang$^3$}
\address{$^1$ University of Washington\quad
$^2$ Taiyuan University of Technology\\
$^3$ Shanghai Jiao Tong University}

\begin{document}
%
\maketitle

\input{src/0-abs}
\input{src/1-intro}

\input{src/2-related}
\input{src/dataset}

\input{src/3-method}
\input{src/4-experiments}
\input{src/5-conclusion}

{
    \bibliographystyle{IEEEbib}
    \bibliography{strings,refs}
}
\end{document}

%% file: src/0-abs.tex
\begin{abstract}
Detecting lane lines from sensors is becoming an increasingly significant part of autonomous driving systems. However, less development has been made on high-definition lane-level mapping based on aerial images, which could automatically build and update offline maps for auto-driving systems. To this end, our work focuses on extracting fine-level detailed lane lines together with their topological structures. This task is challenging since it requires large amounts of data covering different lane types, terrain and regions. In this paper, we introduce for the first time a large-scale aerial image dataset built for lane detection, with high-quality polyline lane annotations on high-resolution images of around 80 kilometers of road. Moreover, we developed a baseline deep learning lane detection method from aerial images, called AerialLaneNet, consisting of two stages. The first stage is to produce coarse-grained results at point level, and the second stage exploits the coarse-grained results and feature to perform the vertex-matching task, producing fine-grained lanes with topology. The experiments show our approach achieves significant improvement compared with the state-of-the-art methods on our new dataset. Our code and new dataset are available at \href{https://github.com/Jiawei-Yao0812/AerialLaneNet}{https://github.com/Jiawei-Yao0812/AerialLaneNet}.
\end{abstract}


%% file: src/1-intro.tex
\vspace{-0.6cm}

\section{Introduction}
The rapid development of intelligent technology~\cite{li2023explicit, liu2023prompt, li2023style, liu2023dap} has led to heavy application of HD mapping techniques in intelligent transportation systems, particularly in the context of autonomous driving. The perception of the environment surrounding the autonomous vehicle is the most fundamental and important technology~\cite{munir2018autonomous, yao2023depthssc, yao2023ndc}, including many challenging tasks~\cite{yao2023geometry}, like lane detection, object detection~\cite{yao2023dynamicbev} and depth estimation~\cite{yao2023improving}. Among them, lane detection is the basic and essential task, whose detection results may affect driving performance and personal safety.

Recently, some researchers introduced deep-learning based method to build road maps from sensor data. Generally, previous methods focus more on performing online lane detection from vehicle's egocentric sensor data and use for planning \cite{liao2022maptr,jiang2023vad}, or building a coarse-level map of the road network or curb map \cite{li2022hdmapnet,xu2022rngdet}. However, building detailed lane maps offline from top-view aerial image data is feasible but not under-studied, and the accuracy and effectiveness of such high-definition mapping are still significantly underperformed. There is a lack of a benchmark and baseline for offline lane recognition from aerial images task. 

Previous lane detection datasets such as SDLane \cite{jin2022eigenlanes}, ONCE-3DLanes \cite{yan2022once}, and OpenLane \cite{chen2022persformer} are all designed for lane detection from a vehicle perspective. In this paper, we introduce a new benchmark for lane detection using aerial images. This benchmark covers various areas and includes different lane types with high-quality labeling and high-resolution images. The dataset contains 7,763 images and over 150,000 lanes covering different lane standards, terrain and regions, providing a comprehensive resource for researchers in this field.

\begin{figure*}[t]
  \centering
\includegraphics[width=0.7\linewidth]{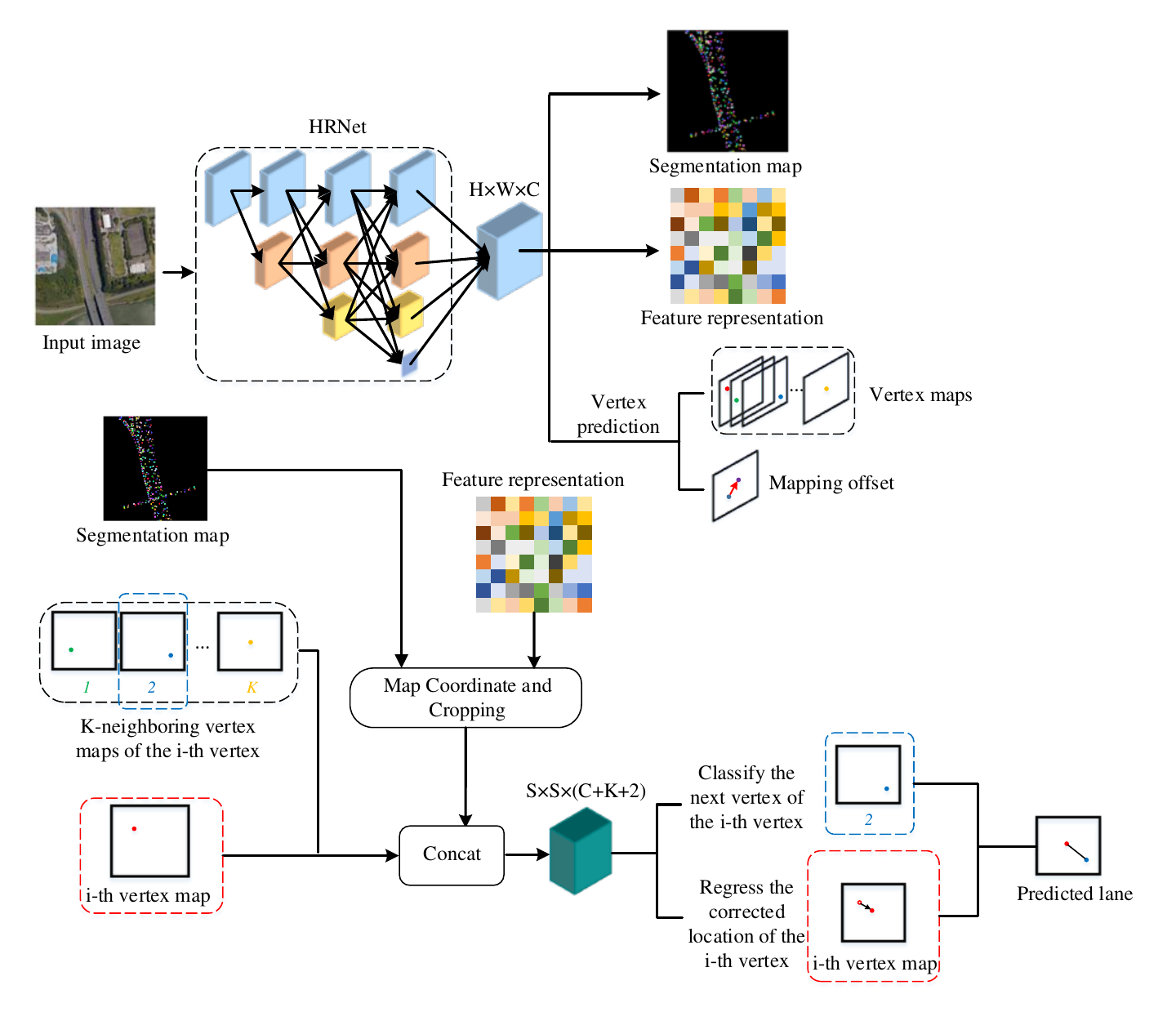}
    \vspace{-0.5cm}
  \caption{Overview of our proposed method AerialLaneNet composed of the first stage: predicting segmentation map, feature map and vertex maps. The second stage takes previous stage as input and predicts each of the next vertex to link the vertices as polyline.}
  \label{fig:overview}
  \vspace{-0.4cm}
\end{figure*}

We also propose a baseline method of building accurate lane maps offline from aerial images. Specifically, we developed a two-stage cascaded lane detection approach called AerialLaneNet, in which the first stage is responsible for producing coarse-grained results and features, and the second stage fully exploits the coarse-grained results and feature maps to perform the vertex matching task and produce fine-grained results at the instance level. The main contribution of this work can be summarized as follows: (1) We present the first-ever benchmark for lane detection from aerial images covering different areas and various lane types with high-quality labeling and high-resolution images. This benchmark can facilitate research in this field. (2) We propose a novel cascaded lane detection approach using aerial images, called AerialLaneNet, which consists of two cascaded stages refining the results from point level to instance level. (3) Comprehensive experiments demonstrated that our method has outstanding performance in comparison with state-of-the-art methods.

%% file: src/dataset.tex
\vspace{-0.4cm}
\section{AErial Lane (AEL) Dataset}
\begin{figure}[ht]
  \centering
\includegraphics[width=1\columnwidth]{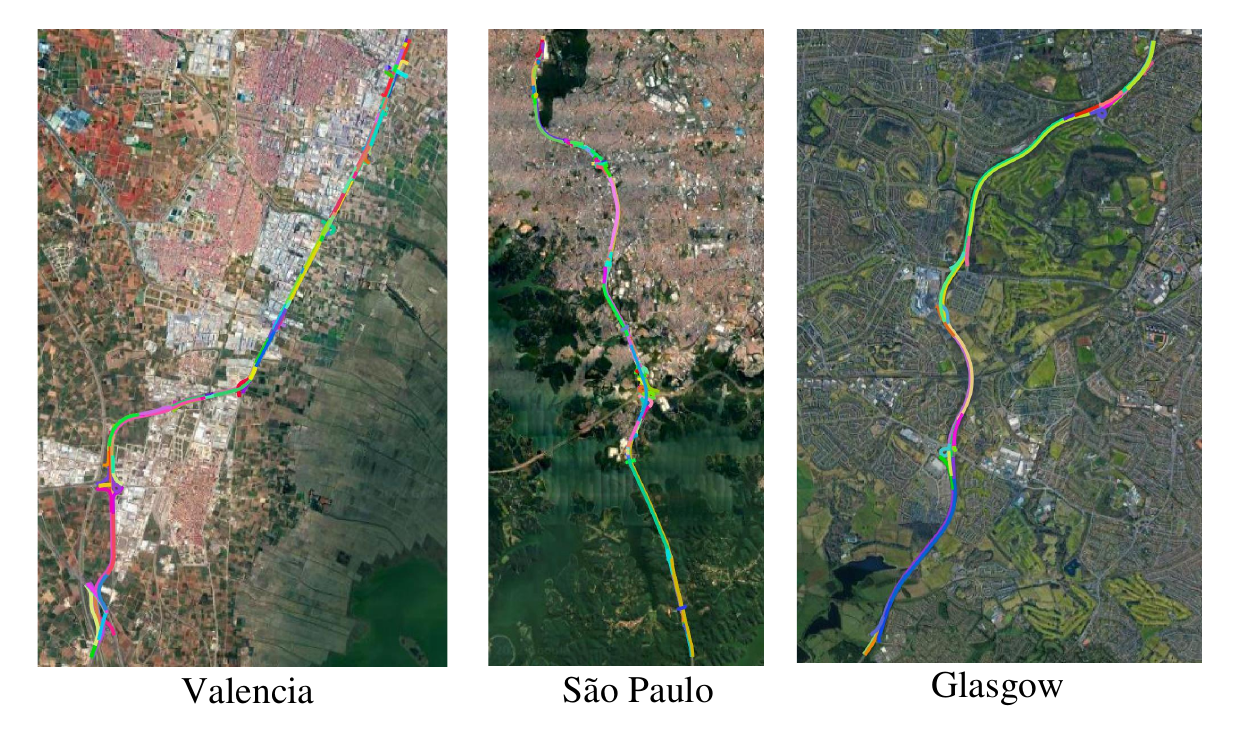}
    \vspace{-1cm}
  \caption{ Three of the annotated region in AEL dataset. Our dataset comprises images from various regions, backgrounds, lane types, topologies, and colors, making the lane detection task challenging. }
  \label{fig:datasets}
  \vspace{-0.4cm}
\end{figure}

\subsection{Overview}

Our work aims to build a benchmark and baseline for detecting vectorized lane lines from high-resolution aerial images. Compared to previous techniques such as detecting dynamic lane maps from ego-vehicle view and road network or curb detection techniques, our task generates accurate lanes as polylines as offline maps for autonomous driving systems to leverage in planning and navigation. Therefore, we propose the AErial Lane (AEL) Dataset, which is the first aerial image dataset with lane-level annotations to our knowledge. The source of images is from Google Map service \cite{map} and we exploit QGIS \cite{qgis} to label the lane lines. We will introduce in this section the specification and statistics for the AEL dataset. 

\vspace{-0.5cm}
\subsection{Annotation Specification}
We directly labeled polylines as the lane lines and the vertices are stored in the form of latitude and longitude coordinates, which makes the dataset easier to share and update. In order to connect vertices into polylines, the direction of the edge is the same in each polyline. In labeling the dataset, the criteria for placing vertex is that the polylines connected by adjacent vertex should approximate the lane line as much as possible, given that the lane lines are actually curves. 

To record the attributes of each lane line in detail, we identify 3 attributes of lanes: single or double line, white or yellow line, solid or dash line. Apart from lane attributes, we record the road\_id that each lane line belongs to, and also assign each lane an id. After annotating the dataset, we export image patches together with the corresponding lane annotation JSON file, the lane mask image from QGIS. We exported images at various heights at random region, and the resolution of every image is $1280\times1280$.

\vspace{-0.2cm}
\begin{table}[ht]
\centering

\begin{tabular}{p{1.5cm}|C{1.5cm}|C{1.8cm}|C{2.2cm}} 
Region & Number of Lanes & Number of Vertices & Total Length (KM) \\ \hline
Cairo & 240 & 2495 & 3.721 \\ \hline
Aucamvile & 130 & 4534 & 17.815 \\ \hline
San Paulo & 779 & 14811 & 26.563 \\ \hline
Nevada & 29 & 1134 & 6.414 \\ \hline
Gopeng & 13 & 1011 & 4.179 \\ \hline
Glasgow & 157 & 4428 & 9.203 \\ \hline
Valencia & 434 & 7116 & 12.044 \\ \hline
\end{tabular}
\caption{The statistics of the AEL dataset, containing the number of lanes, number of vertices, and total road length of each region.}
\vspace{-0.5cm}
\label{tab:stat}
\end{table}
\vspace{-0.2cm}

\subsection{Dataset Statistics}
Firstly, we chose 11 regions, and each region consists of a road between 3 and 27 kilometers long with various backgrounds and terrain such as desert by the coastline in Dubai, urban area of Valencia and forest region in Perak State. Table~\ref{tab:stat} shows the statistics about the annotated lanes with respect to different regions. 

 We then export from QGIS 7,763 images together with over 150,000 lane lines in the form of pixel coordinates for training and evaluating the neural network. Our split ratio of training set:validation set:test set is 7:2:1. Figure~\ref{fig:datasets} demonstrates a few sample images from our proposed dataset. 

\begin{table*}[ht]
\centering
\resizebox{0.8\textwidth}{16mm}{
\begin{tabular}{cccccccccccc}
\hline 
Methods & \multicolumn{3}{c}{
 Precision $\uparrow$} & &  \multicolumn{3}{c}{Recall $\uparrow$} & &  \multicolumn{3}{c}{F1-score $\uparrow$} \\
\cline {2-4}  \cline{6-8} \cline{10-12} & 2.0 & 5.0 & 10.0 & & 2.0 & 5.0 & 10.0 & & 2.0 & 5.0 & 10.0 \\
\hline Naive baseline& 0.607 & 0.890& 0.928 & & 0.505 & 0.736 & 0.768 & &0.533 & 0.778 & 0.811 \\
RoadTracer~\cite{bastani2018roadtracer}& 0.391 & 0.707 & 0.791 & & 0.416 & 0.743 & 0.821 && 0.533 & 0.778 & 0.811 \\
VecRoad~\cite{tan2020vecroad} & 0.461 & 0.769 & 0.854 & & 0.459 & 0.752 & 0.830 && 0.458 & 0.756 & 0.837\\
DAGMapper~\cite{homayounfar2019dagmapper}& 0.407 & 0.751 & 0.868 & & 0.353 & 0.649 & 0.747& &0.371&0.684&0.787\\
iCurb~\cite{xu2021icurb}& 0.550 & 0.833 & 0.890 & & 0.538 & 0.815 & 0.873&&0.542&0.820&0.910 \\
AerialLaneNet & \textbf{0.692} & \textbf{0.925} & \textbf{0.941} & & \textbf{0.613} & \textbf{0.879} & \textbf{0.920} && \textbf{0.613} & \textbf{0.876} & \textbf{0.964}\\
\hline
\end{tabular}}
\caption{\textbf{Quantitative comparsion} against other methods designed for extracting vectorized map elements.}
\label{tab:comparison}
\end{table*}

%% file: src/3-method.tex
\section{The Proposed Baseline: AerialLaneNet}

Figure~\ref{fig:overview} illustrates overall architecture of proposed method, which consists of two stages. The first stage is designed to extract feature representation, and predict the lane segmentation map and vertex maps. In the second stage, each vertex is matched with its next vertex based on the segmentation map, feature representation and vertex maps from stage one. 

For the segmentation branch in the first stage, we adopt HRNet~\cite{wang2020deep} as the backbone network to provide high-resolution representation for image segmentation. With the lane segmentation and high-resolution features at hand, our approach further achieves instance-level lane lines in two steps. Firstly, the feature representation learns lane position information from the segmentation task and is used to predict vertex maps. Next, we will introduce in detail vertex matching using the results produced in the first stage (feature representation, segmentation map, and vertex maps) to obtain instance-level lane lines. 

\subsection{Vertex Prediction}
We predict vertex maps $\hat{Y} \in [0,1] $ with the size $H \times W \times C$, where $H$ and $W$ denote height and width of vertex maps,  respectively, and $C$ represents the max vertex number, which is set to the max vertex number of an image in the dataset. Specifically, $\hat{Y}_{x,y,c} = 1$ represents a detected vertex, and $\hat{Y}_{x,y,c} = 0$  denotes the background.

The vertex prediction process is similar to \cite{law2018cornernet}. For each ground truth $p \in R^2$  of vertex c, there is only one positive value, and the others are zeros. We first define a low-resolution representation $\overline{p} = \lfloor \overline{p} / R \rfloor$  with the downsampling stride $R = 4$ , and then map each vertex to a Gaussian heatmap $Y_{xyc}$. Also, when two Gaussian functions have overlapping areas, we take the maximum value of the heatmaps instead of the average. In this way, the close relationship to peak points can be preserved. The training object of vertex prediction is similar to focal loss $L_{det}$ ~\cite{lin2017focal} .

In the next section, we will introduce vertex matching to find the next vertex for each vertex.

\subsection{Vertex Matching}
In the second stage, we perform a vertex matching task to determine whether a pair of vertexes are from the same lane line. This task is finally fed into two branches: classification and regression, which are designed to match vertexes for each lane line and predict the corrected location for each vertex, respectively. The overall process is illustrated in Figure~\ref{fig:overview}.

For vertex $v_i$, we search its top-K neighboring vertexes according to Euclidean distance from $v_i$ to other vertexes. Then we aggregate location representation and feature maps into a vector, which contains abundant vertex location and high-resolution semantic features. Specifically, we represent the segmentation map as $\hat{Y}_{seg}$ , $T_{v_i} = \{m_{v_{ik}}\}_{k=1}^K$  represents top-K neighboring vertex maps of vertex $v_i$,  $f_{v_i} \in R ^ {H \times W \times (C + K + 2)}$ is the aggregated vector, and $F \in R^{H \times W \times C}$  denotes the feature representation producing from the backbone. The $f_{v_i}$ is generated by concatenating $\hat{Y}_{seg}$, $T_{v_i}$ and $F$ together with vertex map of $v_i$ on the channel dimension. To decrease the computational budget, we crop the $f_{v_i}$ by a square with size $S \times S$, whose center is at the vertex $v_i$. $\bar{f}_{v_i}$  representing the cropped aggregated vector. 

The $\bar{f}_{v_i}$ is finally fed into two branches: classification and regression, which are targeted to classify the next vertex of $v_i$ from K neighboring vertexes and regress the corrected location of $v_i$, respectively. The classification head consists of a linear projection layer followed by a softmax function. We use the cross-entropy loss as the classification loss $L_{cls}$ between true class map $t_{v_i}$ and predicted confidence map $\hat{t}_{v_i}$. 

The regression branch outputs are the location $d(v_i) = (d_x(v_i), d_y(v_i))$  for vertex $v_i$ . $d_x(v_i)$ and $d_y(v_i)$ denotes the prediction of the corrected coordinate of vertex $v_i$. The ground truth location of vertex $v_i$ is represented as $u(v_i) = (u_x(v_i), u_y(v_i))$. We use smooth $L1$ loss as the location regression loss $L_{reg}$. 

Finally, we employ a multi-task training strategy to minimize the objective loss, which comprises five parts: segmentation loss, vertex prediction loss, remapping offset loss, classification loss, and location regression loss. We formulate the objective loss as:
\begin{equation}
L=L_{seg}+L_{det}+\lambda_1 L_{c l s}+\lambda_2 L_{r e g}
\end{equation}
where $\lambda_1$ and $\lambda_2$ are the regularization parameters. We empirically set them as 0.1 and 0.01 in our experiment.

%% file: src/4-experiments.tex
\vspace{-0.2cm}
\section{Experiments}

\subsection{Implementation Details}

Our AerialLaneNet takes as input $1280 \times 1280$ resolution patches randomly cropped from the raw images with random flips. The network is trained with Adam optimizer and cosine annealing schedule with the initial learning rate of $1 e-3$, a momentum of $0.9$ and train the network for 100 epochs. which we drop by a factor of 5 at epoch 40 and 60. We adopt a batch size of 4 for each round. Training the network took around 16 hours. To train the Inception network that reasons about whether a connection in the graph exists. For the segmentation and feature extraction stage, we adopt HRNet~\cite{wang2020deep} as the backbone, and for next vertex prediction we exploit a ResNet-50 as the classifier. 

\begin{table}[ht]
\centering
\begin{tabular}{l|ccc}
\hline K & F1-Score$_{class}$ & MSE$_{position}$ & Runtime$_{class}$ \\
\hline
5 & 80.2 & 10.2 & 0.03 \\
10 & 81.7 & 8.8 & 0.05  \\
20 & 83.8 & 7.4 & 0.10\\
40 & 84.0 & 5.2 & 0.16 \\
\hline
\end{tabular}
\caption{\textbf{Ablation} on K number of vertex candidate. }
\label{tab:ablation}
\end{table}
\vspace{-0.6cm}

\subsection{Results and Comparisons}

\begin{figure}[h]
    \centering
    \includegraphics[width=1\columnwidth]{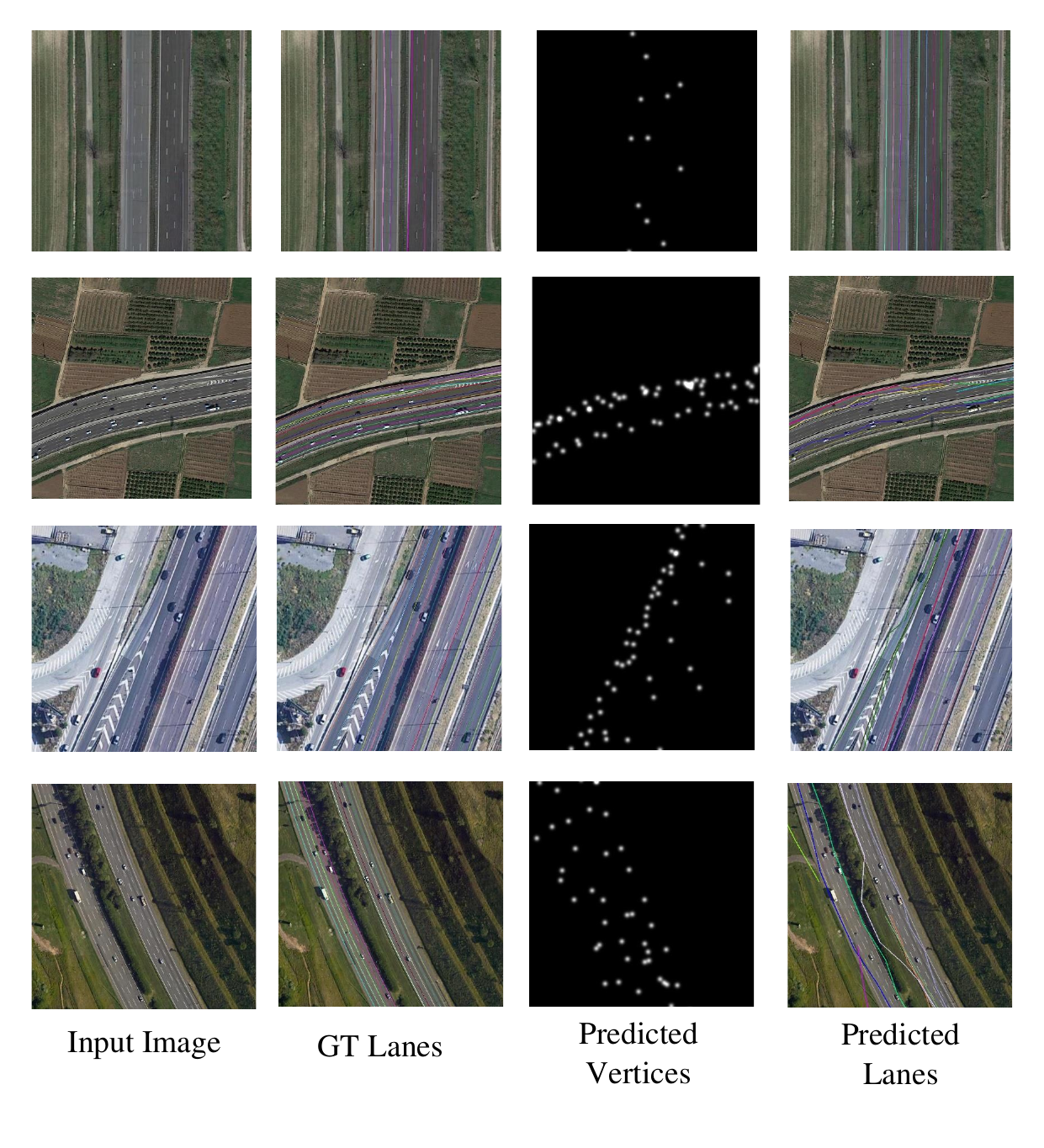}
    \vspace{-0.8cm}
    \caption{The ground truth and baseline experimental results. Fig.3 (a) is the image patch, and Fig.3 (b) is the lane mask generated from polyline annotation. Our dataset consists of high-resolution images, with typical lane widths larger than 5 pixels and minimal occlusion, making lane detection achievable. }
    \label{fig:results}
    \vspace{-0.8cm}
\end{figure}

Table~\ref{tab:comparison} presents the performance of AerialLaneNet outperforms all other methods by a considerable margin and our method excel in Precision, Recall and F1-score under different thresholds, demonstrating the effectiveness and robustness of our proposed architecture. We also provide qualitative evaluation in the form of visualization results. Figure~\ref{fig:results} shows the original image, ground truth lanes and vertices and our prediction results. Our method connected most of the vertices correctly and restored the lane topology correctly from vertices, thus successfully predicting the main structure of lanes. However, due to the shift of vertices location prediction, our predicted lanes sometimes warp compared to the ground truth.

We conduct an ablation study on the number of candidates in the next vertex prediction process. We test several K values, and the results and shown in Table~\ref{tab:ablation}. We found that with the increasing number of K, the final MSE of the next vertex position regression significantly drop down, indicating a better prediction on next vertex position. The F1-Score of classifying the next vertex increases as well. However, the runtime increases linearly, due to the extra burden of classification.

\vspace{-0.4cm}

%% file: src/5-conclusion.tex
\section{Conclusion}

In this paper, we propose a dataset consisting of aerial images with various backgrounds and fine-grained lane types, which will be made publicly available upon publication of this paper. Furthermore, we introduce a baseline method, AerialLaneNet, for detecting lane lines using aerial images. Extensive experiments confirm that the proposed method consistently outperforms state-of-the-art methods, and our study can be utilized in online automatic mapping, enabling autonomous driving and intelligent transportation applications.